\RequirePackage{scrlfile}
\documentclass{article}

\PassOptionsToPackage{numbers, compress}{natbib}

\PreventPackageFromLoading{subfig}

\usepackage[final]{neurips_2019_ml4ps}

\usepackage[utf8]{inputenc} 
\usepackage[T1]{fontenc}    
\usepackage{hyperref}       
\usepackage{graphicx}
\usepackage{svg}
\usepackage{wrapfig}
\usepackage{subcaption}
\usepackage{url}            
\usepackage{booktabs}       
\usepackage{amsfonts}       
\usepackage{nicefrac}       
\usepackage{microtype}      
\usepackage{color}
\usepackage{textcomp}
\usepackage{float}

\newcommand{\forecasttime}[1]{t={#1}\textrm{h}}

\title{Predicting Weather Uncertainty with Deep Convnets}

\author{%
  Peter Gr\"{o}nquist \\
  ETH Z\"{u}rich \\
  \texttt{petergro@student.ethz.ch}
  \And
  Tal Ben-Nun \\
  ETH Z\"{u}rich \\
  \texttt{tal.bennun@inf.ethz.ch}
  \And
  Nikoli Dryden \\
  ETH Z\"{u}rich \\
  \texttt{nikoli.dryden@inf.ethz.ch}
  \And
  Peter Dueben \\
  ECMWF \\
  \texttt{peter.dueben@ecmwf.int}
  \AND
  Luca Lavarini \\
  ETH Z\"{u}rich \\
  \texttt{lucalav@student.ethz.ch}
  \And
  Shigang Li \\
  ETH Z\"{u}rich \\
  \texttt{shigang.li@inf.ethz.ch}
  \And
  Torsten Hoefler \\
  ETH Z\"{u}rich \\
  \texttt{htor@inf.ethz.ch}
}

\begin{document}

\maketitle

\begin{abstract}
Modern weather forecast models perform uncertainty quantification using ensemble prediction systems, which collect nonparametric statistics based on multiple perturbed simulations. To provide accurate estimation, dozens of such computationally intensive simulations must be run. We show that deep neural networks can be used on a small set of numerical weather simulations to estimate the spread of a weather forecast, significantly reducing computational cost. To train the system, we both modify the 3D U-Net architecture and explore models that incorporate temporal data. Our models serve as a starting point to improve uncertainty quantification in current real-time weather forecasting systems, which is vital for predicting extreme events.
\end{abstract}

\section{Introduction}

\begin{wrapfigure}{r}{0.5\textwidth}
\vspace{-1em}
  \begin{center}
    \includegraphics[width=\linewidth]{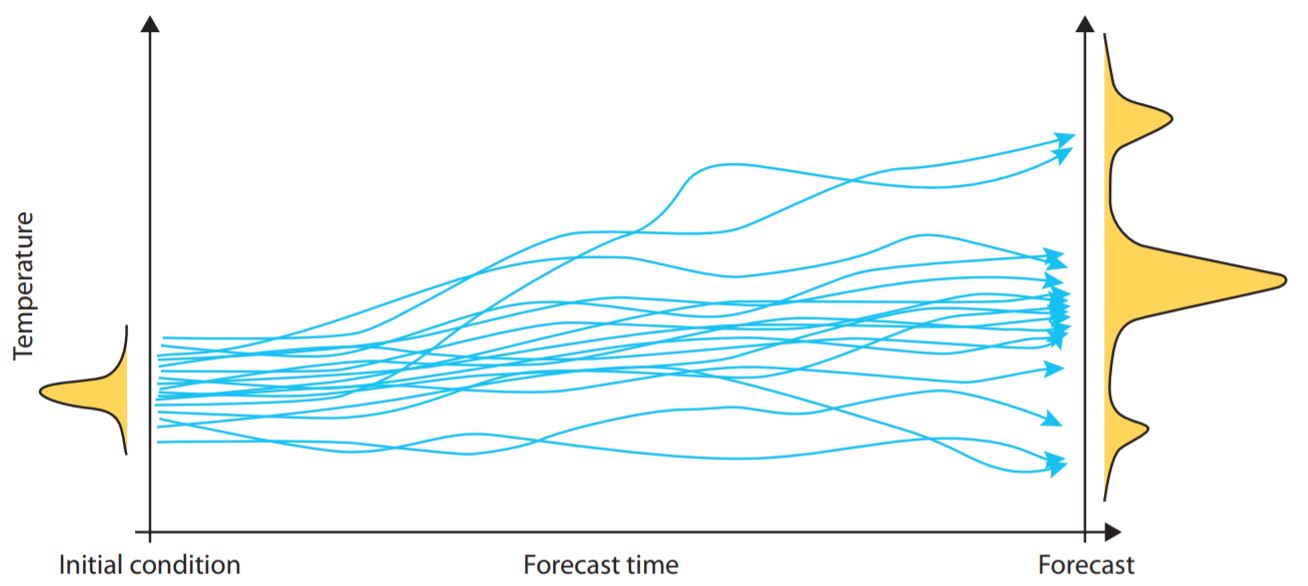}
    \caption{ECMWF Ensemble Prediction System.}
    \label{fig:ens_forecast}
    \end{center}
    \vspace{-1em}
\end{wrapfigure} 
Weather forecasting is of critical importance for many areas of life, from civil safety to food security. Uncertainty quantification of weather forecasts is essential due to the chaotic dynamics of weather and the exponential growth of forecast errors in time. When a tropical cyclone approaches a coastline, it is important to know both the most likely position of landfall and the probability that it will hit other locations. Thus, numerical weather predictions (NWP) need to provide reliable probability distributions for their predictions. To this end, weather centers, such as the European Centre for Medium-Range Weather Forecasts (ECMWF), typically run an ensemble of unique weather forecasts (\emph{trajectories}) in parallel, to estimate a probability distribution for several parameters, e.g. temperature ~\cite{palmer2000predicting,ecmwfensembles} (Figure~\ref{fig:ens_forecast}). Each ensemble member is perturbed, and the difference in the resulting predictions, measured by their standard deviation (\emph{spread}), can be used to identify the uncertainty of a high-resolution forecast. Modeling such distributions is important for predictions of extreme weather events, and requires as many as fifty ensemble members~\cite{leutbecher2018ensemble}. Running these ensemble simulations with many members at a global scale dramatically increases the computational demands of weather forecast models.

In an effort to speed up parts of NWP, recent works apply machine learning techniques to weather and climate models~\cite{rolnick2019tackling}. Most works revolve around augmenting physical models for improved accuracy~\cite{mcgovern2017using,dueben2018challenges,scher2018toward,mudigonda2017segmenting,racah2017extremeweather,kurth2018exascale}, or for short-term predictions such as precipitation nowcasting~\cite{xingjian2015convolutional,shi2017deep,heye2017precipitation}. However, there has been relatively little work on using machine learning for uncertainty quantification of multi-parameter forecasts and global data. Previous works focus instead on producing a single scalar value~\cite{scher2018predicting} or station-specific information~\cite{rasp18ens} on local regions.


In this paper, we present preliminary work exploring 3D CNN architectures for predicting the spread of ensemble forecasts using as few as one simulation with our initial parameters. Our forecasts come from the ECMWF ensemble forecasting system, which is based on the Integrated Forecasting System (IFS)~\cite{ifs}. We consider several novel architectures for exploiting spatial and temporal effects, such as affine convolutions, as well as study existing architectures, such as CNN-LSTMs~\cite{xingjian2015convolutional}. Once trained, our models are able to approximate the mean and spread of a large ensemble of models accurately and with low computational cost.
 
\section{Data}

For our preliminary exploration, we use the ERA5 dataset~\cite{era5}, as it is similar to the data used by production NWP and is publicly available. The dataset is produced by the ECMWF, and currently consists of weather data reanalysis from 1979 to the present. It includes an ensemble of nine perturbed trajectories and a single unperturbed (control) trajectory, of which we explore different subsets. The available data was mapped to a latitude and longitude grid with a 0.5 degree resolution and contains 37 pressure levels. We use temperature prediction as an initial target. Based on previous exploratory work~\cite{merenti2002analysing}, we select a subset of Initial Parameters (IP) that have an influence on temperature: zonal and meridional wind, geopotential, temperature, relative humidity, and the fraction of cloud cover. 
\begin{wrapfigure}{r}{0.6\textwidth}
  \centering
  \includegraphics[scale=0.3]{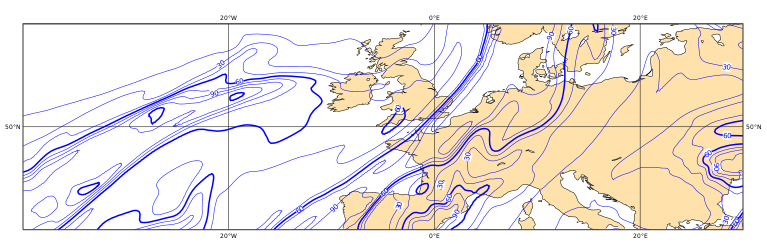}
  \caption{Example of relative humidity in our selected area, at 150 hPa from longitudes -40° to 30° and latitudes 40° to 60°, plotted using ECMWF’s Magics library~\cite{magics}.}
  \label{fig:area_plot}
\vspace{-1em}
\end{wrapfigure} 

We use a subset of the data that includes Europe and parts of the Atlantic Ocean (Figure~\ref{fig:area_plot}), with a window of 40 latitude by 136 longitude points. We choose seven pressure levels, including 500 hPa (which is in the middle of the atmosphere) and 850 hPa (close to the surface), which can be used to identify warm and cold fronts due to limited daily temperature variations. For the temporal dimension, we use measurements from 0600 and 1800 UTC, with forecasts made for three ($t=3$h) and six ($t=6$h) hours into the future.

Our subset of the ERA5 data is available in the GRIB format~\cite{grib} for the years 2000-2011, from which we extract our region of interest and standardize the data for each pressure level and parameter. We save the data as single-precision floats (due to TensorFlow limitations) in chronological order and in the correct spatial distribution in NumPy arrays~\cite{van2011numpy}. We then convert the data to the TFRecord~\cite{abadi2016tensorflow} format with one year per file. We use the most recent years (2010, 2011) as a test set; of the remaining data, we randomly select 80\% for training and 20\% for validation. We shuffle data to ensure that training and validation datasets are drawn from a variety of times and that seasons are not observed in sequence.

We also consider a second dataset that we call ENS10. ENS10 is based on ECMWF re-forecasts that perform 10-member ensemble predictions two times per week for the last twenty years~\cite{ens10}. All trajectories are perturbed. Re-forecasts are very similar to the operational 51-member ensemble used by the ECMWF but run at coarser resolution of 0.5 degrees. This allows us to gain additional insight into what our model's expected performance on the 51-member ensemble would be, as this data is not available to us.

\section{Model and Methods}

\begin{figure}
  \centering
  \includegraphics[scale=0.35]{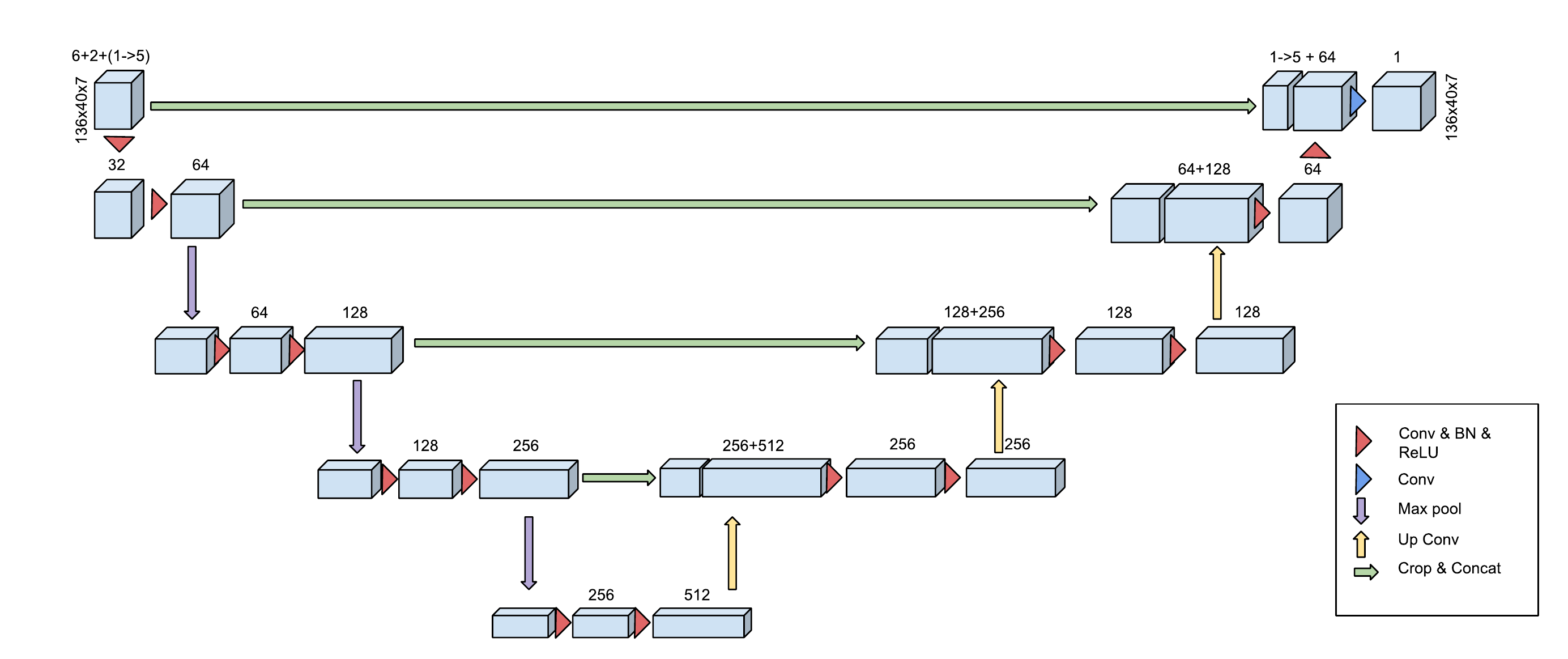}
  \caption{Our baseline model, adapted from 3D U-Net~\cite{cciccek20163d}.}
  \label{fig:basic_model}
\end{figure}


To comprehensively explore this problem, we consider three aspects: integrating spatial effects from multiple pressure levels and resolutions, predicting temperature for all pressure levels at once; learning from temporal trends by including previous forecasting timesteps, predicting temperature one pressure level at a time; and efficient implementations.

Our baseline model is adapted from the 3D U-Net~\cite{cciccek20163d} DNN, which contains residual connections to preserve spatial information (Figure~\ref{fig:basic_model}). We concatenate all parameters and trajectories channel-wise to form each input sample. We also considered a DeepLab v3+ model~\cite{chen2018encoder} with a ResNet-50 backbone~\cite{he2016deep}, but observed that it performed 9.6\% worse than the results we report here for our baseline model on temporal data.

Due to the chaotic nature of weather, small perturbations have a large influence on weather patterns, especially for the long-term forecasts we are trying to predict. We speculate that it is harder to extract information from these smaller spatial patterns in our data when they are downsampled through convolutions and max pooling compared to natural images. This limits the benefit of deeper networks.

\begin{wrapfigure}{r}{0.6\textwidth}
  \centering
  \begin{subfigure}[t]{0.47\linewidth}
  \includegraphics[height=0.6in,page=1]{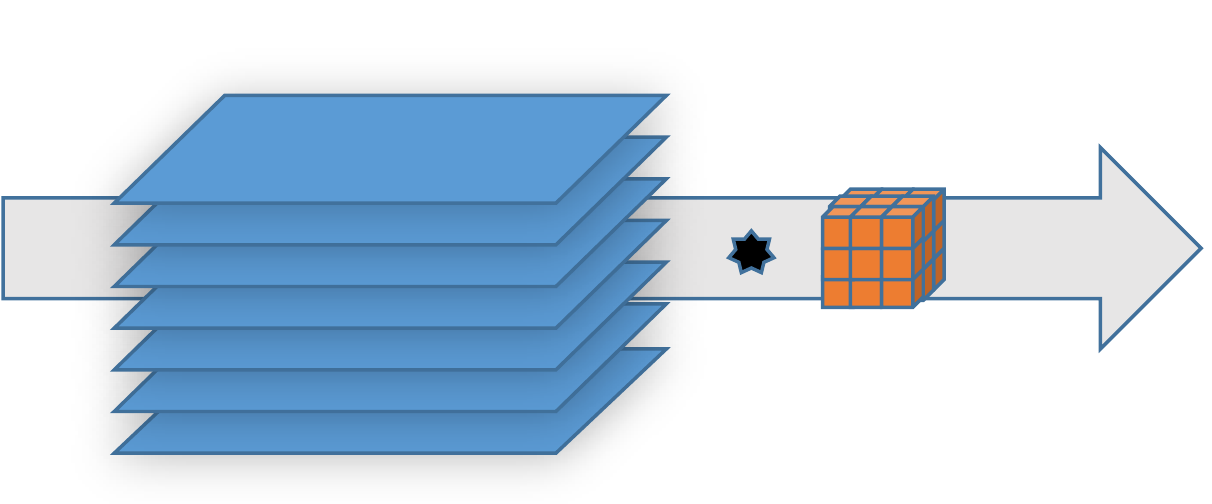}
  \caption{Standard}
  \end{subfigure}\hfil
  \begin{subfigure}[t]{0.47\linewidth}
  \includegraphics[height=0.65in,page=3]{figures/convs.pdf}
  \caption{Full}
  \end{subfigure}
  \begin{subfigure}[t]{0.47\linewidth}
  \includegraphics[width=\linewidth,height=0.6in,page=2]{figures/convs.pdf}
  \caption{Affine}
  \end{subfigure}\hfil
  \begin{subfigure}[t]{0.47\linewidth}
  \includegraphics[height=0.65in,page=4]{figures/convs.pdf}
  \caption{Separable}
  \end{subfigure}
  \caption{Different convolutions for exploring spatial effects.}
  \label{fig:aff}
  \vspace{-1em}
\end{wrapfigure}

\textbf{Spatial effects.} It is challenging to apply 3D convolution directly to our data, because grid points are not uniformly spaced. This is most evident in the pressure levels, where distances increase exponentially with respect to geopotential height. For large maps, Gaussian grids also hamper the use of simple convolution windows. We consider several approaches to address this, as depicted in Figure~\ref{fig:aff}. Weight sharing on each pressure level separately (``Full'') enables more flexibility to learn different representations, but significantly increases the number of parameters. More conservatively, we can introduce learned point-wise affine transforms per pressure level after each convolution (``Affine''). As a compromise between parameters and representational power, we also consider horizontal 2D convolutions followed by vertical 1D convolutions (``Separable'').

\textbf{Temporal effects.} To explore temporal trends, we look at the data of the spread of all ten trajectories at times $\forecasttime{0}$, $\forecasttime{3}$ and $\forecasttime{6}$. Our goal is to see if our models can learn the temporal progression of the spread. As a first step, we only use the spread at $\forecasttime{0}$ and $\forecasttime{3}$ as input. Subsequently we also explore the additional concatenation of Initial Parameters (IP) from the unperturbed trajectory. Our basic method is inspired by the work on precipitation nowcasting~\cite{shi2017deep}. We also evaluate CNN-LSTMs~\cite{xingjian2015convolutional} as an efficient method that also retains spatial information. However, we find that with our limited time slices, simply treating time sequences as additional channels in U-Net and ResNet structures offers the best performance. Fully understanding the impact of temporal data on our predictions requires additional study with more timesteps.

\textbf{Distributed training.} To efficiently train on the global high-resolution data, as necessary in production NWP, it is crucial to employ distributed training. In particular, the memory requirements of training on such data can easily exceed GPU memory. We currently leverage distributed data-parallelism and propose to further mitigate this problem with a combination of pipeline parallelism for network depth~\cite{huang2018gpipe,li2018pipe} and model parallelism for high-resolution data~\cite{dryden2019improving,dryden2019channel}.

\begin{figure}[t]
    \centering
    \captionsetup[subfigure]{justification=centering}
        \begin{subfigure}[t]{0.5\textwidth}
            \centering
            \includegraphics[height=1.3in]{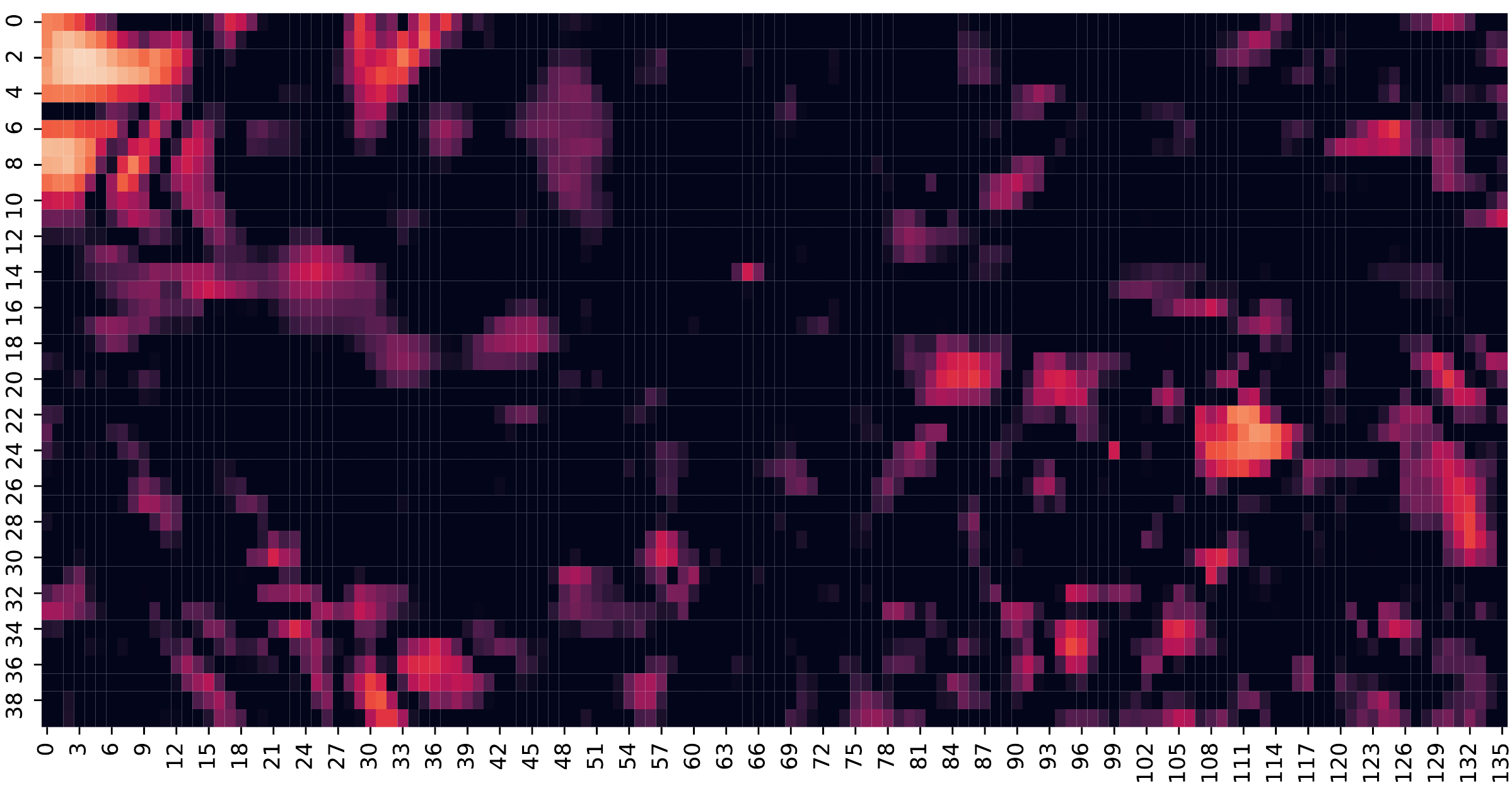}
            \caption{Linear regression from spread at $\forecasttime{0}$.\\
            RMSE = 0.1346}
        \end{subfigure}%
        ~
        \begin{subfigure}[t]{0.5\textwidth}
            \centering
            \includegraphics[height=1.3in]{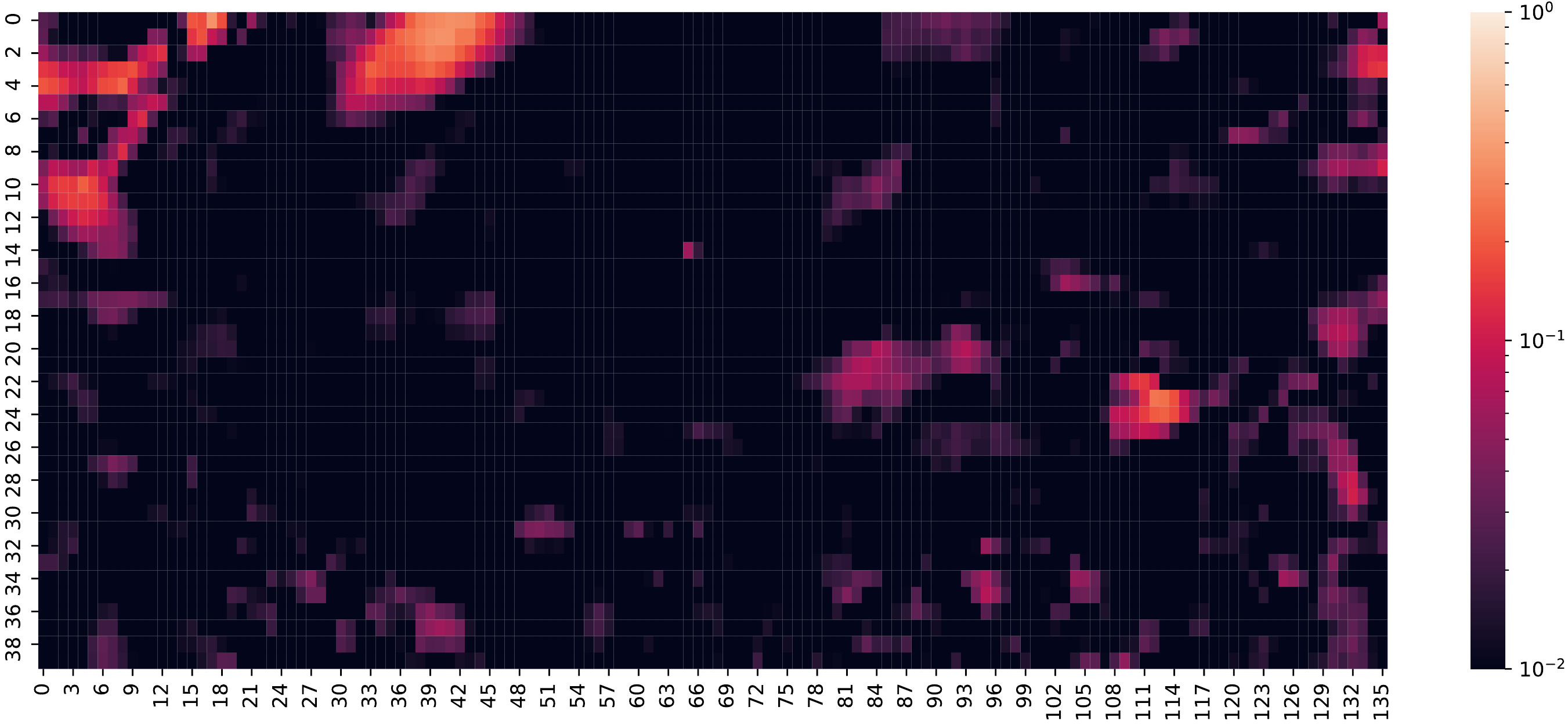}
            \caption{Baseline model prediction.\\
            RMSE = 0.1060}
        \end{subfigure}
        \caption{Logarithmic heatmap of squared difference towards full ensemble spread at $\forecasttime{6}$ (850 hPa, temperature in K). Axes are our selected longitude and latitude.}
        \label{fig:MSE_PLOT}
\end{figure}

\section{Evaluation}

\begin{figure}
  \centering
  \includegraphics[width=0.9\linewidth]{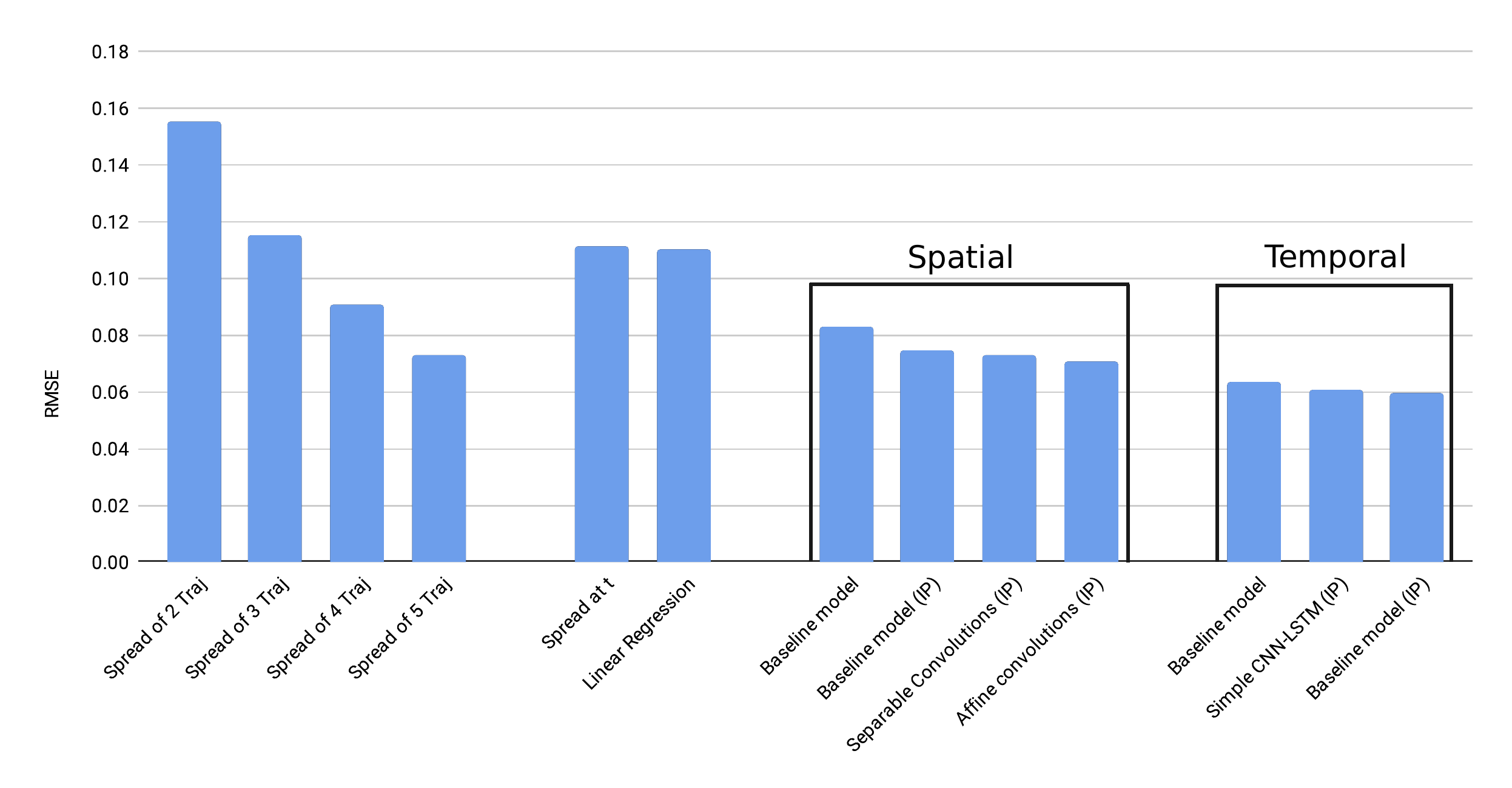}
  \caption{ERA5 test RMSE compared to the full ensemble spread.}
  \label{fig:RMSE}
\end{figure}

We compare the initial results of our models using the root mean squared error (RMSE) as the optimization target, showing the final test values in Figure~\ref{fig:RMSE}. As an algorithmic comparison, we use a linear regression on the full ensemble spread at time $\forecasttime{0}$, which has about the same error as the spread of three trajectories. Our baseline model, using only one unperturbed trajectory, provides a better spread estimation than using four perturbed trajectories. Additional changes to our model that include spatial effects put us on equal footing with five trajectories (half of what is available in ERA5). We further improve upon this by incorporating temporal data and predicting $\forecasttime{6}$.

We also evaluate the impact of concatenating our Initial Parameters (IP) to the input data in Figure~\ref{fig:RMSE}. While it does show an improvement, it is minor compared to the impact of model architecture.

As the RMSE does not incorporate spatial coherency, we visualize the predictions of our baseline model to better understand them (Figure~\ref{fig:MSE_PLOT}). We observe that the model places greater importance on larger shifting spread regions (upper left corner) while neglecting some of the lower spread regions. This shows promise for detecting extreme weather events.

\begin{wrapfigure}{r}{0.4\linewidth}
  \centering
  \includegraphics[width=\linewidth]{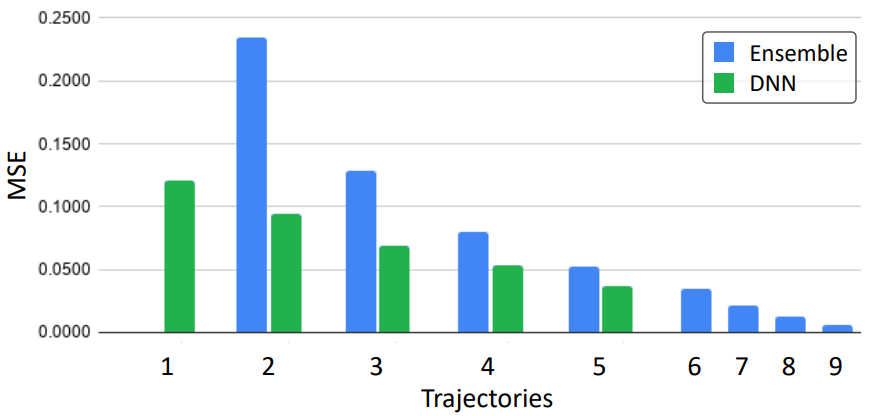}
  \caption{ENS10 final validation loss (baseline model, global).}
  \label{fig:ENS10VLOSS}
\end{wrapfigure}

Finally, we evaluate our baseline model on the ENS10 dataset on a global scale and observe similar promising results (Figure~\ref{fig:ENS10VLOSS}). Our model successfully approximates larger ensembles using only a few input trajectories. We observe diminishing returns (MSE towards the 10-member ensemble spread) as we increase the number of trajectories used. We theorize that this is due to the small number of ensemble members in the dataset: Five trajectories is half of the ensemble. Additionally, we do not observe any benefit from our temporal models. We hypothesize this is because the temporal grid in ENS10 (24 hour intervals) is coarser than in ERA5 (3 hour intervals), generating weather pattern changes that are significantly harder to predict.

It is too early to provide a comprehensive cost/benefit comparison between conventional ensemble predictions and those post-processed with deep learning. However, current NWP models use significant time on state-of-the-art supercomputers. ECMWF's 51-member ensembles run twice per day for one hour on a Cray XC40 supercomputer with more than 4 PetaFLOPs peak performance. These calculations use and predict roughly ten times more pressure levels and diverse parameters for 15-day forecasts. In comparison, for our models to predict seven pressure levels for one parameter and 2-day forecasts takes about 15 ms on one Nvidia V100 GPU, which would be performed after the trajectory calculations. Thus, even when scaling up to full-resolution production predictions, our models will be considerably faster than running an ensemble trajectory. Further, there remain many optimization opportunities in the training and inference pipelines.

\section{Discussion}

We have demonstrated promising preliminary results on using CNNs to approximate ensembles for NWP with significantly reduced computational requirements. This was possible due to the use of techniques such as affine convolution to incorporate spatial information and CNN-LSTMs to incorporate temporal information. This serves as an important first step toward integrating deep learning into production NWP pipelines and improving real-time weather forecasting. By using our method to reduce the number of ensemble members, we can both reduce the compute requirements and improve the forecast product for simulations that are constrained by the time required to run current NWP models. In the future, we plan to evaluate these models on larger and higher-resolution datasets containing global information. This introduces additional challenges, such as nonuniform grid points and higher computational and memory costs for training.

\section*{Acknowledgments}
This project has received funding from the European Research Council (ERC) under the European Union's Horizon 2020 programme (grant agreement DAPP, No. 678880).

\bibliographystyle{IEEEtran}
\bibliography{refs}

\end{document}